\documentclass[runningheads]{llncs}

\usepackage[utf8]{inputenc} 
\usepackage[T1]{fontenc}    
\usepackage{hyperref}       
\usepackage{url}            
\usepackage{booktabs}       
\usepackage{amsfonts}       
\usepackage{nicefrac}       
\usepackage{microtype}      
\usepackage[square,sort,comma,numbers]{natbib}
\usepackage{graphicx}
\usepackage{enumitem}
\usepackage{subfigure}
\usepackage{wrapfig}
\usepackage{adjustbox}

\newcommand{\ourmodel}{\textsc{CorD-CPD }}
\newcommand{\ourmodelshort}{\textsc{CorD-CPD}}
\usepackage{bm}
\usepackage{amssymb}
\usepackage{amsmath}
\usepackage{multirow}
\usepackage{multicol}
\usepackage{comment}
\usepackage{dsfont}
\usepackage{bm}
\usepackage{xcolor}

\newcommand{\black}[1]{\vspace{3pt}\noindent\textbf{#1}}

\title{Correlation-aware Change-point Detection \\
        via Graph Neural Networks}

\author{Ruohong Zhang*\inst{1} \and
Yu Hao*\inst{1} \and 
Donghan Yu \inst{1} \and
Wei-Cheng Chang \inst{1} \and
Guokun Lai \inst{1} \and
Yiming Yang \inst{1}}
\authorrunning{R. Zhang et al.}
\institute{Carnegie Mellon University, 5000 Forbes Ave, Pittsburgh, PA 15213,
\email{\{ruohongz,yuhao2,dyu2,wchang2,guokun,yiming\}@cs.cmu.edu}}

\begin{document}

\maketitle

\begin{abstract}
Change-point detection (CPD) aims to detect abrupt changes over time series data. Intuitively, effective CPD over multivariate time series should require explicit modeling of the dependencies across input variables.  However, existing CPD methods either ignore the dependency structures entirely or rely on the (unrealistic) assumption that the correlation structures are static over time.
In this paper, we propose a Correlation-aware Dynamics Model for CPD, which explicitly models the correlation structure and dynamics of variables by incorporating graph neural networks into an encoder-decoder framework. Extensive experiments on synthetic and real-world datasets demonstrate the advantageous performance of the proposed model on CPD tasks over strong baselines, as well as its ability to classify the change-points as correlation changes or independent changes. \footnote{ $\ast$ indicates equal contribution of authors. \\ Code is available at https://github.com/RifleZhang/CORD\_CPD}
\end{abstract}

\section{Introduction}
 
\begin{wrapfigure}{R}{0.5\textwidth}
     \centering
     \includegraphics[width=\linewidth]{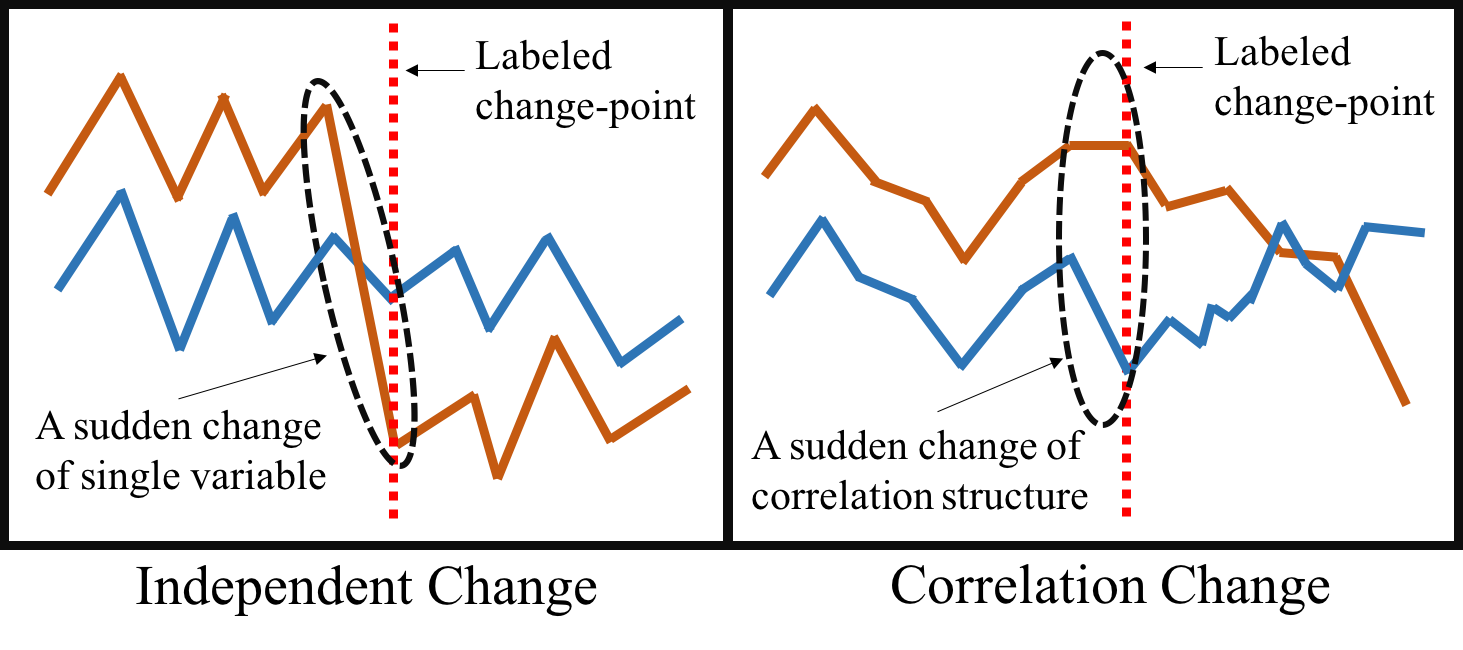}
     \caption{(Left) an independent change of one variable and (Right) a correlation change between two variables. The red vertical line is the labeled change-point.}
     \label{fig:intro}
     \vspace{-15pt}
\end{wrapfigure}

Change-point detection (CPD) aims to detect abrupt property changes over time series data. In this study, change-points are detected through the changes of \textit{dynamics} and \textit{correlation} of variables. Dynamics refers to the physical property that determines a variable's modus operandi and correlation describes the interactions between variables. 
Previous CPD methods~\cite{zhang2010detecting, montanez2015inertial} model dynamics by parametric distributions like Hidden Markov Models (HMM), but they don't explicitly capture the correlation information. Other works capture static correlation structures in the multivariate time series~\cite{kipf2018neural}, but they can't detect any correlation changes.
We propose a \textbf{Cor}relation-aware \textbf{D}ynamics Model for \textbf{C}hange-\textbf{p}oint \textbf{D}etection (\ourmodelshort) which incorporates graph neural networks into an encoder-decoder framework to explicitly model both changeable correlation structure and variable dynamics. We refer to the changes of correlation structure as \textbf{correlation changes} and the changes of variable dynamics as \textbf{independent changes}, as shown in Fig. \ref{fig:intro}.

Our model is capable of distinguishing the two types of changes, which could have a broader impact on decision-making. In financial markets, traders use pair trading strategy to profit from correlated stocks, such as Apple and Samsung (both are phone sellers), which share similar dips and highs. News about Apple expanding markets may independently raise its price without breaking its correlation with Samsung. However, news about Apple building self-driving cars will break its correlation with Samsung, and establish new correlations with automobile companies. While both of them are change-points, the former is an independent change of variables and the latter is a correlation change between variables. Knowing the type of change can guide financial experts to choose trading strategies properly.

Our contributions can be summarized as follows:

\begin{itemize}
    \item We propose \ourmodel to capture both changeable correlation structure and variable dynamics.
    \item Our \ourmodel classifies the change-points as correlation changes or independent changes, and ensembles them for robust CPD.
    \item Experiment on synthetic and real datasets demonstrates that our model can bring enhanced interpretability and improved performance in CPD tasks. 
\end{itemize}

\section{Method for CPD}
\label{sec:background}
A multivariate time series is denoted by $\mathbf{x} \in \mathbf{R}^{T\times N\times M}$, where $T$ is the time steps, $N$ is the number of variables and $M$ is the number of features for each variable.
We study the CPD problem in a retrospective setting and assume there is one change-point per $\mathbf{x}=\{\mathbf{x}^j\}_{j=1}^{T}$.
The change-point at time step $t$ satisfies:
\setlength{\abovedisplayskip}{2pt}
\setlength{\belowdisplayskip}{2pt}
\begin{align*}
    \{ \mathbf{x}^1, \mathbf{x}^2, \ldots, \mathbf{x}^{t-1} \} &\sim \mathbb{P} \\
    \{ \mathbf{x}^{t}, \mathbf{x}^{t+1}, \ldots, \mathbf{x}^{T} \} &\sim \mathbb{Q}
\end{align*}
Where $\mathbb{P}$ and $\mathbb{Q}$ denotes two different distributions. We attribute this difference to a correlation change (of the correlation structure), an independent change (of variable dynamics), or a mixture of both.

\black{Correlation Change} corresponds to the change of the correlation structure of multivariate time series, which is modeled by correlation matrices $\mathbf{A} \in \mathbb{R}^{T\times N \times N}$. 
At each time step, the pairwise interaction between variables ($A^t_{ij}$) is represented as a continuous value between $0$ and $1$, indicating how much they are correlated.
The correlation change score $\mathbf{s}_r$ is calculated by the $L_1$ distance between two neighboring correlation matrices:
\begin{equation}
    s_{r}^t = \| \mathbf{A}^t - \mathbf{A}^{t-1}\|_1 \text{, } t > 1 
\end{equation}  

\black{Independent Change} corresponds to the change of the variable dynamics. Given the current values of time series (and the extracted correlation matrices), if the dynamics rule is followed, the expected values of the future time steps predicted by our model will be close to the observed values; Otherwise, the difference will be large. This difference is used as the independent change score $\mathbf{s}_d$. 
Formally, we use the Mean Squared Error (MSE) as a metric to compare the expected values $\hat{\mathbf{w}}^{t+1}=\{\hat{\mathbf{x}}^i\}_{i=t+1}^{t+k}$ with the observed values $\mathbf{w}^{t+1}=\{\mathbf{x}^i\}_{i=t+1}^{t+k}$ over a window of size $k$.
\begin{equation}
    s_d^t = \operatorname{MSE}( \hat{\mathbf{w}}^{t}, \mathbf{w}^{t}) \text{, } t > 1
\end{equation}
Note that if only a correlation change takes place, the expected value $\hat{\mathbf{w}}^{t}$ should not be different from the observed value $\mathbf{w}^{t}$, since we model a conditional probability $P(\mathbf{x}^t|\mathbf{x}^{<t},\mathbf{A})$ and any correlation change will be factored in.

\black{Ensemble of Change-point Scores} aims to combine the correlation change with the independent change, because in real world applications, change-points could be resulted from a mixture of both. A simple way to ensemble them (for $\mathbf{s}_{en}$) is to sum the normalized scores of $\mathbf{s}_r$ and $\mathbf{s}_d$:
\begin{align}
    &\mathbf{s}_{en} = \operatorname{Norm}(\mathbf{s}_{r}) + \operatorname{Norm}(\mathbf{s}_{d}) \\
    &\operatorname{Norm}(\mathbf{s}) = \frac{\mathbf{s} - u_s}{\sigma_s} 
\end{align}
where $u_s$ and $\sigma_s$ are mean and standard deviation of score $\mathbf{s}$.

In order to use our CPD methods above, we need to model correlation matrices and to be able to predict a future window of time steps based on the extracted correlation. We will introduce our \ourmodel in the next section. 

\section{Correlation-aware Dynamics Model}
\label{sec:method}

The \ourmodel has an encoder for correlation extraction and a decoder for variable dynamics. Given a time series $\mathbf{x}$, the encoder models a distribution of correlation matrix $q_\phi(\mathbf{A}^t | \mathbf{x})$ for each time step $t$, and by factorization,
\begin{equation}
    q_\phi(\mathbf{A}|\mathbf{x}) = \prod_{t=1}^T q_\phi({\mathbf{A}^t|\mathbf{x}})
\end{equation}
The decoder models a distribution of time steps $p_\theta(\mathbf{x}|\mathbf{A})$ auto-regressively,
\begin{equation}
    p_\theta(\mathbf{x}|\mathbf{A}) = \prod_{t=1}^T p_\theta(\mathbf{x}^{t}|\mathbf{x}^{<t}, \mathbf{A}^{t-1})
\end{equation} 
\noindent The objective function maximizes the log likelihood,
\begin{equation}
    \mathcal{L}_{obj} = \mathbb{E}_{q_\phi({\mathbf{A|X}})}[\log p_\theta(\mathbf{x}|\mathbf{A})]
\end{equation}

\begin{figure*}[t!]
     \centering
     \includegraphics[width=\linewidth]{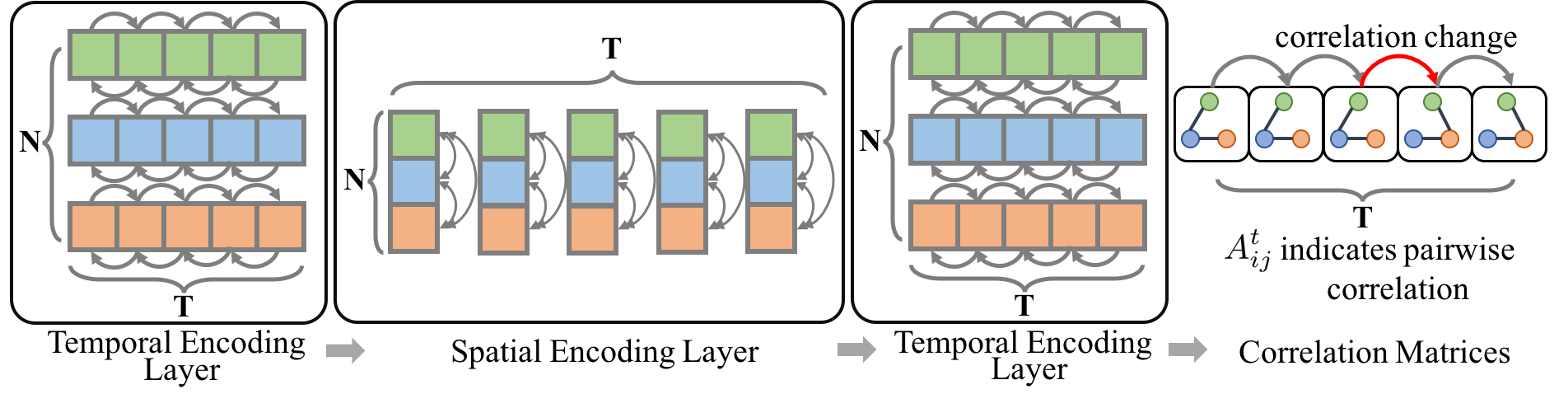}
     \caption{\ourmodel Encoder: the encoder extracts correlation matrices from multivariate time series. The temporal encoding layer captures time dependent features, and the spatial encoding layer models relational features between variables.}
     \label{fig:enc}
\end{figure*}

\subsection{Correlation Encoder}
The encoder infers a correlation matrix $\mathbf{A}^t$ at each time step, which depends on both temporal features and variable interactions. 
To leverage both sources, we propose Temporal Encoding Layers (TEL) to extract features across time steps and Spatial Encoding Layers (SEL) to extract features from variable interactions. As shown in Fig \ref{fig:enc}, the two types of layers are alternatively applied to progressively incorporate temporal and correlation features into latent embeddings. Practically, we found $2$ TEL and $1$ SEL is enough for our tasks.

For each layer, let $\mathbf{h} \in \mathbb{R}^{T\times N \times K}$ denote the input and let $\tilde{\mathbf{h}} \in \mathbb{R}^{T\times N \times K'}$ denote the output, where $T$ is the time steps, $N$ is the number of variables, and $K, K'$ are the number of input and output features respectively. The input to the initial layer is the multivariate time series data $\mathbf{x} \in \mathbb{R}^{T\times N \times M}$. The posterior distribution of the correlation matrix is modeled by
\begin{align}
    \label{eq-encoder}
    q_\phi(A^t_{ij}|\mathbf{x}) &= \operatorname{Softmax}(\operatorname{Linear}([\tilde{\mathbf{h}}_{(f)i}^t;  \tilde{\mathbf{h}}_{(f)j}^t]) \\
    \tilde{\mathbf{h}}_{(f)} &= \operatorname{TEL_2}(\operatorname{SEL}(\operatorname{TEL_1}(\mathbf{x})))
\end{align}
where $[\cdot;\cdot]$ is the concatenation operator and $\tilde{\mathbf{h}}_{(f)}$ is the embedding of the final layer. As an additional trick, we apply Gumbel-Softmax \cite{jang2016categorical} to enforce sparse connections in correlation matrices in order to reduce noise.

\black{Temporal Encoding Layer (TEL)} leverages information across $T$ time steps (independently for each variable). 
For a fixed variable $i$, let $\mathbf{h}_i = \{\mathbf{h}^t_i\}_{t=1}^{t=N}$ denote the embeddings of that variable at all time steps.
We offer two implementations of TEL with different neural architectures: RNN\textsubscript{TEL} and Trans\textsubscript{TEL}.
\paragraph{RNN\textsubscript{TEL}} is a bidirectional GRU network~\cite{cho2014learning}:
\begin{align}
    \overrightarrow{\mathbf{h}_i}^t &= \overrightarrow{\text{GRU}}(\overrightarrow{\mathbf{h}_i}^{t-1}, \mathbf{h}^t_i) \\
    \overleftarrow{\mathbf{h}_i}^t &= \overleftarrow{\text{GRU}}(\overleftarrow{\mathbf{h}_i}^{t+1}, \mathbf{h}^t_i) \\
    \tilde{\mathbf{h}}_i^t &= [\overrightarrow{\mathbf{h}_i}^t, \overleftarrow{\mathbf{h}_i}^t]
\end{align}
where $\overrightarrow{\mathbf{h}_i}^t, \overleftarrow{\mathbf{h}_i}^t$ are intermediate representation from forward and backward GRU. The output $\tilde{\mathbf{h}}^t$ is a concatenation of embeddings from both directions.

\paragraph{Trans\textsubscript{TEL}} uses the Transformer model~\cite{vaswani2017attention} with self-attention to capture temporal dependencies. For the self-attention layer, the input is transformed into query matrices $\mathbf{Q}_i^t = \mathbf{h}_i^t \mathbf{W}_Q$, key matrices $\mathbf{K}_i^t = \mathbf{h}_i^t\mathbf{W}_K$ and value matrices $\mathbf{V}_i^t = \mathbf{h}_i^t\mathbf{W}_V$. Here $\mathbf{W}_Q, \mathbf{W}_K, \mathbf{W}_V$ are learnable parameters. Finally, the dot-product attention is a weighted sum of value vectors:
\begin{equation}
\tilde{\mathbf{h}}_i^t = \operatorname{softmax}\left(\frac{\mathbf{Q} \mathbf{K}^{T}}{\sqrt{d_{k}}} \right)\cdot \mathbf{V}
\end{equation}
where $d_k$ is the size of hidden dimension. Similar to \cite{vaswani2017attention}, we use residual connection, layer normalization and positional encoding for Trans\textsubscript{TEL}.

\black{Spatial Encoding Layer (SEL)} leverages the information between the $N$ variables (independently at each time step) via graph neural networks (GNN) \cite{kipf2016semi}. 
For a fixed time step $t$, let $\mathbf{h}^t = \{\mathbf{h}^t_i\}_{i=1}^{N}$ denote the embeddings all variables at time $t$. The output is obtained by
\begin{equation}
      \tilde{\mathbf{h}}^t = \operatorname{GNN}(\{\mathbf{h}^t_i\}_{i=1}^{N}) \\
\end{equation}
where a GNN module is implemented by the feature aggregation and combination operations:
\begin{align}
 \mathbf{e}_{ij} &= f_e([\mathbf{h}_i^t;  \mathbf{h}_j^t])  \label{eq:agg} \\
\tilde{\mathbf{h}}_j &= f_v(\mathbf{h}_{j} + \sum_{i\neq j} \mathbf{e}_{ij}) \label{eq:com}
\end{align}
where Eq. \ref{eq:agg} aggregates features between neighboring nodes and Eq. \ref{eq:com} combines those features by a summation. $f_e(\cdot)$ and $f_v(\cdot)$ are non-linear neural networks for which we provide two implementations: GNN\textsubscript{SEL} and Trans\textsubscript{SEL}.

GNN\textsubscript{SEL} is implemented by a multilayer perceptron (MLP) and Trans\textsubscript{SEL} is implemented by the Transformer model. Compared with MLP, Transformer has could be advantageous for spatial encoding because of well-designed self-attention, residual connection and layer normalization. The positional encoding layer is removed from Transformer because the variables are order invariant.

\subsection{Dynamics Decoder}

\begin{wrapfigure}{R}{0.4\textwidth}
    \vspace{-25pt}
     \centering
     \includegraphics[width=\linewidth]{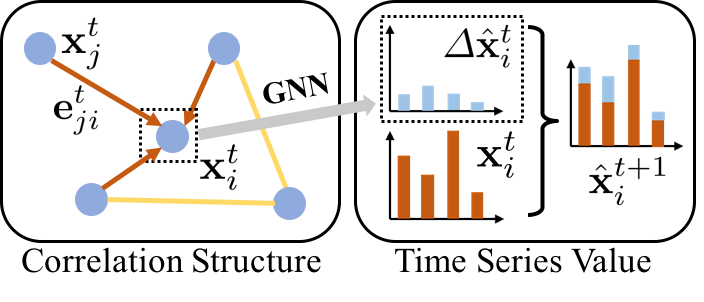}
     \caption{\ourmodel decoder: Given a correlation matrix, the decoder predicts the change of future steps.}
     \label{fig:dec}
     \vspace{-25pt}
\end{wrapfigure}
At a high level, the decoder learns the dynamics of variables by predicting the future time steps to be as close as the observed values. Instead of predicting the value of $\hat{\mathbf{x}}^{t+1}_i$ directly, we predict the change $\Delta \hat{\mathbf{x}}^{t}_{i} = \hat{\mathbf{x}}^{t+1}_i - \mathbf{x}^t_i$ as shown in Fig. \ref{fig:dec}. 

Since the prediction has to factor in the correlation between variables, we also need GNN to incorporate correlation matrices into feature embeddings. 
Again, the feature aggregation and combination operations are performed on the input $\mathbf{x}^t$,
\begin{align}
\mathbf{e}^t_{ji} &= \mathbf{A}^t_{ji}g_e([\mathbf{x}_j^t ; \mathbf{x}_i^t]) \\
\tilde{\mathbf{h}}^t_{i} &= g_v(\mathbf{x}^t_i + \sum_{j\neq i} \mathbf{e}^t_{ji})
\end{align}
where the functions $g_e(.)$ and $g_v(.)$ are MLPs. We model
$\Delta \hat{\mathbf{x}}^t_i = g_{\text{out}}(\tilde{\mathbf{h}}^{\le t}_i)$,
where $g_{\text{out}}(\tilde{\mathbf{h}}^{\le t}_i)$ can be $\operatorname{MLP}(\tilde{\mathbf{h}}^{t}_i)$ or $\operatorname{RNN}(\tilde{\mathbf{h}}^{\le t}_i)$ depending on the application.
Together, $\hat{\mathbf{x}}^{t+1}_i = \mathbf{x}^t_i + \Delta \hat{\mathbf{x}}^{t}_i
    = \mathbf{x}^t_i + g_{\text{out}}(\tilde{\mathbf{h}}^{\le t}_i)$.
    
\noindent The log likelihood of density $p_\theta(\mathbf{x} | \mathbf{A})$ can be expressed as:
\begin{align}
    \log p_\theta(\mathbf{x} | \mathbf{A}) &= \sum_{t=1}^T \log p_\theta(\mathbf{x}^{t}|\mathbf{x}^{<t}, \mathbf{A}^{t-1}) \\
     &= \sum_{i}\sum_{t=1}^T \log \mathcal{N}(\mathbf{x}_i^{t}| \hat{\mathbf{x}}_i^{t}, \sigma^2\mathbf{I}) \\
     &\propto -\sum_i \sum_{t=2}^T \frac{\|\mathbf{x}^t_i- \hat{\mathbf{x}}^t_i\|_2^2}{2\sigma^2} \label{eq:obj}
\end{align}
Maximizing Eq. \ref{eq:obj} is equivalent to minimizing $\mathcal{L}_{obj} = \sum_i \sum_{t=2}^T \frac{\|\mathbf{x}^t_i- \hat{\mathbf{x}}^t_i\|_2^2}{2\sigma^2}$.

Since change-points are sparse in time series data, we introduce an additional regularization to ensure the smoothness of correlation matrix:
\begin{equation}
    \mathcal{L}_{smooth} = \frac{1}{T-1} \sum_{t=2}^T \|\mathbf{A}^t - \mathbf{A}^{t-1}\|_2^2 
\end{equation}
Finally, the loss function is 
$\mathcal{L} = \mathcal{L}_{obj} + \lambda \mathcal{L}_{smooth}$, 
where $\lambda$ controls the relative strength of smoothness regularization.

\section{Experiment with Physics Simulations}
\label{sec:synthetic}
\subsection{Particle-spring Change-point Dataset}
We developed a dataset with a simulated physical particle-spring system. The system contains $N=5$ particles that move in a rectangular space. Some randomly selected pairs (out of the $10$ pairs in total) of particles are connected by invisible springs. The motion of particles are determined by the laws of physics such as Newton's law, Hooke's law, and Markov property. The trajectories of length $T=100$ of the particles are recorded as the multivariate time series data. Each variable has $M=4$ features: location $l_x, l_y$ and speed $v_x, v_y$.

While the physical system is similar to the one in \cite{kipf2018neural}, we additionally design $3$ types of change-points by perturbing the location, speed, and connection at a random time step between $[25, 75]$:
\begin{itemize}[noitemsep,topsep=0pt,parsep=0pt,partopsep=0pt]
    \item \textbf{location}: A perturbation to the current location sampled from $\mathcal{N}(0, 0.1)$, where the range of the location is $[-5, 5]$.
    \item \textbf{speed}: A perturbation to the current speed by sampled from $\mathcal{N}(0, 0.02)$, where range of the speed is $[-1, 1]$.
    \item \textbf{connection}: re-sample connections and ensure that at least $5$ out of $10$ pairs of connections are changed.
\end{itemize}
The change of location or speed (both are dynamics) belongs to the independent change, and the change of connection (a type of correlation) belongs to the correlation change. Since the change-point is either a correlation change or an independent change, we are able to test the ability of our model to classify them. 

We generate $500$ time series for each type of change and mix them together (totally $1500$ time series) as training data. For validation and testing data, we generate $100$ time series for each type of change and evaluate on them separately. Our model is unsupervised, so the validation set is only used for hyperparameter tuning. In real world datasets, human labeled change-points are scarce in quantity, which usually results in large variance in evaluation. As a remedy, our synthetic data can be generated in a large amount to reduce such a variance in testing.

\subsection{Evaluation Metric and Baselines}
For quantitative evaluation of CPD performance, we consider two metrics: 

\black{Area-Under-the-Curve (AUC)} of the receiver operating characteristic (ROC) is a metric commonly used in the CPD literature~\cite{chang2019kernel}.

\black{Triangle Utility (TRI)} is a hinge-loss-based metric: $\max(0, 1 - \frac{\|y-l\|}{w})$, where $w=15$ is the margin, $l$ and $y$ are the labeled and predicted change-points.  

\vspace{3pt}
\noindent Both of the metrics range from $[0, 1]$ and higher values indicate better predictions. However, AUC treats the change-point scores at each time step independently, without considering any temporal patterns. TRI considers the distance between the label and the predicted change-point (the one with highest change-point score), but it doesn't measure the quality of predictions at the other time steps. We use both metrics because they complement with each other.

Next, we introduce $6$ baselines of the state-of-the-art statistical and deep learning models:
\begin{itemize}[noitemsep,topsep=0pt,parsep=0pt,partopsep=0pt]
    \item \textbf{ARGP-BODPD}~\cite{saatcci2010gaussian} is Bayesian change-point model that uses auto-regressive Gaussian Process as underlying predictive model.
    \item \textbf{RDR-KCPD}~\cite{liu2013change} uses relative density ratio technique that considers f-divergence as the dissimilarity measure.
    \item \textbf{Mstats-KCPD}~\cite{li2015m} uses kernel maximum mean discrepancy (MMD) as dissimilarity measure on data space.
    \item \textbf{KL-CPD}~\cite{chang2019kernel} uses deep neural models for kernel learning and generative method to learn pseudo anomaly distribution.
    \item \textbf{RNN}~\cite{cho2014learning} is a recurrent neural network baseline to learn variable dynamics from multivariate time series (without modeling correlations). 
    \item \textbf{LSTNet} \cite{lai2018modeling} combines CNN and RNN to learn variable dynamics from long and short-term temporal data (without modeling correlations).
 \end{itemize}

\begin{table*}[ht!]
\begin{center}
\setlength{\tabcolsep}{6pt}
\begin{tabular}{c ||c c | c c | c c}
\hline
\bf \multirow{2}{*}{model} & \multicolumn{2}{c|}{\bf location} & \multicolumn{2}{c|}{\bf speed} & \multicolumn{2}{c}{\bf connection} \\ 
 & AUC & TRI & AUC & TRI & AUC & TRI  \\
\hline
\bf ARGP-BOCPD & 0.5244 & 0.0880 & 0.5231  & 0.0660 & 0.5442  & 0.1287  \\
\bf RDR-KCPD & 0.5095 & 0.0680 & 0.5279 & 0.1093 & 0.5234  & 0.0860 \\
\bf Mstats-KCPD & 0.5380 & 0.0730 & 0.5369  & 0.0727 & 0.5508  & 0.0833 \\
\hline
\bf RNN & 0.5413  & 0.2567 & 0.5381  & 0.2660 & 0.5446  & 0.3047 \\
\bf LSTNet & 0.5817 & 0.3487 & 0.5817  & 0.3460 & 0.5337  & 0.2193 \\
\bf KL-CPD & 0.5247 & 0.1053 & 0.5378 & 0.1352 & 0.5574  & 0.3127 \\
\hline
\bf GNN\textsubscript{SEL}+RNN\textsubscript{TEL} & 0.9864 & 0.9740 & 0.9700  & \bf 0.9320 & \bf 0.9681 & \bf 0.9153 \\
\bf Trans\textsubscript{SEL}+RNN\textsubscript{TEL} & \bf 0.9885 & \bf 0.9773 & \bf 0.9755  & 0.9080 & 0.9469  & 0.9040 \\
\bf GNN\textsubscript{SEL}+Trans\textsubscript{TEL} & 0.9692 & 0.9333 & 0.9609  & 0.8473  & 0.8840 & 0.8527 \\
\hline
\end{tabular}
\end{center}
\caption{\label{tab:cpd} AUC and TRI metrics on synthetic datasets for the prediction of location, speed and connection change. Our \ourmodel (evaluated with $\mathbf{s}_{en}$) has the best performance on both metrics among all the baselines.}
\vspace{-20pt}
\end{table*}

\subsection{Main Results}
Table \ref{tab:cpd} shows the performance of the statistical baselines (first panel), the deep learning baselines (second panel) and our proposed \ourmodel (third panel).

\black{Statistical Baselines} are not as competitive as the other deep learning models among all the types of changes. One explanation is that those models have strong assumption on the parameterization of probability distributions, which may hurt the performance on datasets that demonstrate complicated interactions of variables. The dynamics rule of the physics system can be hardly captured by those methods. 

\black{Deep Learning Baselines} are slightly better than the statistical models, in which the LSTNet has the best performance on location and speed changes. Since LSTNet has a powerful feature extractor for long and short-term temporal data, it is better at learning variable dynamics. However, as correlation plays an important role in the synthetic data, ignoring it will hurt performance in general.

\black{\ourmodelshort} is evaluated on the test data by the ensemble score $\mathbf{s}_{en}$. It has the best performance on both metrics among all the baselines. We didn't include the result of Trans\textsubscript{SEL}$+$Trans\textsubscript{TEL},
because empirically it is harder to converge.
Trans\textsubscript{SEL}$+$RNN\textsubscript{TEL} is the best at detecting the independent changes, while GNN\textsubscript{SEL}$+$RNN\textsubscript{TEL} is the best at detecting the correlation changes. The reason could be that the Transformer models are better at identifying local patterns, while RNNs are more stable at combining features with long term dependencies. Among the three types of changes, the score of connection change is lower than that of the other two, indicating the detection of correlation changes is harder than independent changes.

\begin{table*}[ht!]
\begin{center}
\setlength{\tabcolsep}{4.5pt}
\begin{tabular}{c@{\hskip 0pt} c ||c c|c c|c c}
\hline
\bf \multirow{2}{*}{model} & \multirow{2}{*}{\bf type} & \multicolumn{2}{c|}{\bf location} & \multicolumn{2}{c|}{\bf speed}& \multicolumn{2}{c}{\bf connection} \\ 
 & & AUC & TRI & AUC & TRI & AUC & TRI  \\
\hline
\multirow{2}{*}{\bf GNN\textsubscript{SEL}+RNN\textsubscript{TEL}} & cor & 0.5145 & 0.3153 & 0.5590 & 0.3553 & \bf 0.9649 & \bf 0.9073 \\
 & ind &  \bf 0.9835 & \bf 0.9727 & \bf 0.9587 & \bf 0.9493 & 0.8093 & 0.7320 \\
\hline
\multirow{2}{*}{\bf Trans\textsubscript{SEL}+RNN\textsubscript{TEL}} & cor & 0.4944 & 0.2626 & 0.5463 & 0.3266 & \bf 0.9755 & \bf 0.9273  \\
 & ind & \bf 0.9859 & \bf 0.9720 & \bf 0.9685  & \bf 0.9233 & 0.7774 & 0.6460 \\
\hline
\multirow{2}{*}{\bf GNN\textsubscript{SEL}+Trans\textsubscript{TEL}} & cor & 0.5544 & 0.3467 & 0.5832 & 0.4266 & \bf 0.9098  & \bf 0.8787  \\
 & ind & \bf 0.9855  & \bf 0.9693 & \bf 0.9623  & \bf 0.9133 & 0.7912 & 0.7620  \\
\hline
\end{tabular}
\end{center}
\caption{\label{tab:sep} Our \ourmodel separately computes the scores for correlation change (cor) and independent change (ind). The correlation change score is high on the connection data, while the correlation change score is high on the location and speed data.}
\end{table*}

\subsection{Change-point Type Classification}
Our \ourmodel separately computes change-point scores for correlation change ($\mathbf{s}_r$) and independent change ($\mathbf{s}_d$). We show the ability of our model to separate the two types of changes based on the scores.

\subsubsection{Correlation Vs. Independent Change.}
In Table \ref{tab:sep}, the correlation change (cor) and 
independent change (ind) are separately evaluated. The correlation change scores ($\mathbf{s}_r$) are high on the connection data, while the independent change scores ($\mathbf{s}_d$) are high on location and speed change. This result shows that our system can indeed distinguish the two types of changes.

\paragraph{Location\&speed} The independent changes can be successfully distinguished. For location and speed data, AUC of the independent changes is over $0.97$, close to a perfect detection; AUC of the correlation change is close to $0.5$, nearly a random guess. Therefore, our system doesn't signal a correlation change for location and speed data, but it gives a strong signal of an independent change.

\paragraph{Connection} The correlation changes are harder to be detected, but \ourmodel gives a good estimation. In the connection data, AUC of correlation change are higher than independent change, but the gap was smaller than that in location and speed data. The reason could be that the errors made by encoder are propagated into the decoder, and thus made the forecasting of time series values inaccurate. 

\subsubsection{Classification Method.}
While our model shows a potential to distinguish the two types of changes, we want it to be able to classify them. We propose to use the difference between normalized correlation change score $\mathbf{s}_r$ and independent change score $\mathbf{s}_d$ as an indicator of change-point type, at time $t$:
$$
\operatorname{Norm}(\mathbf{s}_r)^t - \alpha \operatorname{Norm}(\mathbf{s}_d)^t
\begin{cases}
     \ge \tau, & \text{ correlation change} \\
      < \tau, & \text{ independent change}
\end{cases}
$$
where $\alpha=0.75$ is our design choice, and $\tau$ is a threshold to separate the correlation change and the independent change. Moving the value of $\tau$ controls the type I error (False Positive) and the type II error (False Negative). To measure the classification quality by leveraging the error, ROC AUC is a typical solution. 

We classify the change-point types under two settings: with label and without label, according to whether the labeled change-point is provided. 
\paragraph{With Label:} When a labeled change-point is provided by human experts, our model classifies it as either a correlation change or an independent change, whichever dominates. 
\vspace{-8pt}
\paragraph{Without Label:} When the label information is unavailable, our model performs classification from the predicted change-point with the highest $\mathbf{s}_{en}$ score.

\vspace{8pt}
The results are shown in Table \ref{tab:classification}. Our best model Trans\textsubscript{SEL}$+$RNN\textsubscript{TEL} achieves an ROC AUC of $0.979$ (with label) and $0.973$ (without label). This indicates that our model has a strong ability to discriminate the two types of change-points under both settings. GNN\textsubscript{SEL}$+$Trans\textsubscript{TEL} has the worst classification performance, which is consistent to the observation in Table \ref{tab:sep} that it is not good at capturing correlation changes.

In the next experiment, we set $\tau = 0$ and report the classification accuracy on the three data types. As shown in right part of Table \ref{tab:classification}, a high accuracy of $98\%$ on identifying the location and speed change demonstrates that our model can predict the independent changes well. For correlation changes, the Trans\textsubscript{SEL}$+$RNN\textsubscript{TEL} shows the best performance by achieving $93\%$ on supervised setting and $84\%$ on unsupervised setting.

When labeled change-points are not provided, the classification task could be more difficult, because the it relies on the predicted change-points. If a predicted change-point is far from the ground truth, the classification is prone to errors.

\begin{table}[ht!]
    \centering
    \setlength{\tabcolsep}{6pt}
    \begin{tabular}{c || c |c c c}
    \hline
    \bf {model} &  \bf ROC AUC & \bf location & \bf speed & \bf connection \\
    \hline
    With Label & & & & \\
    \hline
     \bf GNN\textsubscript{SEL}+RNN\textsubscript{TEL} & 0.972 & \bf 98\% & \bf 97\% & 68\% \\
     \bf Trans\textsubscript{SEL}+RNN\textsubscript{TEL} & \bf{0.979} & \bf 98\% & 96\% & \bf 93\% \\
     \bf GNN\textsubscript{SEL}+Trans\textsubscript{TEL} & 0.916 & \bf 98\% & 96\% & 87\% \\
    \hline
    Without Label & & & & \\
    \hline
     \bf GNN\textsubscript{SEL}+RNN\textsubscript{TEL} & 0.969 & \bf 98\% & \bf 96\% & 73\% \\
     \bf Trans\textsubscript{SEL}+RNN\textsubscript{TEL} & \bf 0.973 & 96\% & 92\% & \bf 84\% \\
     \bf GNN\textsubscript{SEL}+Trans\textsubscript{TEL} & 0.929 & 91\% & 83\% & 75\% \\
    \hline
    \end{tabular}
    \vspace{10pt}
    \caption{ROC AUC metric demonstrates the ability of our model to separate the two types of change-points. When $\tau=0$, we report the change-point classification accuracy on the $3$ types of data.}
    \label{tab:classification}
    \vspace{-30pt}
\end{table}


\begin{table}[ht!]
\begin{center}
\renewcommand{\arraystretch}{0.9}
\setlength{\tabcolsep}{8pt}
\begin{tabular}{c |c c c}
\hline
\bf {model} &  \bf AUC  & \bf TRI\\
\hline
\bf ARGP-BOCPD & 0.5079  & 0.1773  \\
\bf RDR-KCPD & 0.5633 & 0.1933  \\
\bf Mstats-KCPD & 0.5112 & 0.1480  \\
\hline
 \bf RNN & 0.5540 &  0.2393 \\
 \bf LSTNet & 0.5688 & 0.3145 \\
 \bf KL-CPD & 0.5326 &  0.2102 \\

\hline
 \bf GNN\textsubscript{SEL}+RNN\textsubscript{TEL} & 0.7868 & 0.7574 \\
 \bf Trans\textsubscript{SEL}+RNN\textsubscript{TEL} & 0.7903 & 0.7750 \\
 \bf GNN\textsubscript{SEL}+Trans\textsubscript{TEL} & \bf 0.8277 & \bf 0.8020 \\
\hline
\end{tabular}
\end{center}
\caption{\label{tab:real-data} We report the performance of our \ourmodel on a real-world multivariate time series dataset (PAMAP2). The variables are sensors and the features includes temporatures and 3-D motions. The change-points are transitions between activities.}
\end{table}

\section{Experiments with Physical Activity Monitoring}
\label{sec:real}
In addition to our synthetic dataset, we test our \ourmodel on real-world data: the PAMAP2 Physical Activity Monitoring dataset \cite{reiss2012introducing}. The dataset contains sensor data collected from $9$ subjects performing $18$ different physical activities, such as walking, cycling, playing soccer, etc. Specifically, the variables we consider are $N=3$ Inertial Measurement Units (IMU) on wrist, chest and ankle respectively, measuring $M=10$ features including temperature, 3D acceleration, gyroscope and magnetometer. The change-points are labeled as the transitions between activities. 

To account for the transitions between activities, the independent changes could possibly include the rising of temperature and the correlation changes could be from the switch of different moving patterns between wrist, chest and ankle.

The data was sample every $0.01$ second over totally $10$ hours. In the pre-processing, we down-sample the time-series by $20$ time steps and then slice them into windows of a fixed length $T=100$ steps. Each window contains exactly one transition from range $[25, 75]$. There are totally $184$ multivariate time series with change-points: $150$ of them are used as training, $14$ are used as validation and $20$ are used as testing. 

The results are shown in Table \ref{tab:real-data}. Our \ourmodel achieves the best performance among the $6$ statistical and deep learning baselines. We attribute the enhanced performance to the ability of \ourmodel to better model the two types of changes and to successfully ensemble them. In real life scenarios, a change-point could arise from a mixture of independent change and correlation change. The experiment results show that explicitly modeling both types of changes injects a positive inductive bias during learning, and thus enhances the performance of CPD tasks.

\section{Conclusion}
In this paper, we study the CPD problem on multivariate time series data under the retrospective setting. We propose \ourmodel to explicitly model the correlation structure by incorporating graph neural networks into an encoder-decoder framework. \ourmodel can classify change-points into two types: the correlation change and the independent change. We conduct extensive experiments on physics simulation dataset to demonstrate that \ourmodel can distinguish the two types of change-points. We also test it on the real-word PAMAP2 dataset to show the enhanced performance on CPD over competitive statistical and deep learning baselines.

\appendix
\section{Synthetic Data Demo}
Examples of the synthetic change-point data are shown in Figure \ref{fig:syn_demo}. The trajectories of $5$ particles are displayed before and after a change-point, where the dashed lines represent the expected trajectory if no change-point happened, and the solid lines are the observed trajectory.

\subsection{Location Change}
The location change example is shown in Figure \ref{fig:syn_a}. In this case, we treat the location as multivariate time series, and we observe a small shift of location at and after the change-point. The gap between the expected value and observed value is maintained during the particle movement, but it may vary due to the complicated interactions between those variables.

\subsection{Velocity Change}
The velocity change example is shown in Figure \ref{fig:syn_b}. Compared with the location change, there is no immediate shift in the time series value observed. The gap between the expected value and the observed value tend to become more obvious over time. This is due to the nature of speed such that a small perturbation can cause large difference in location over long time. In order to detect such kind of changes, a window based comparison (of expected values and observed values) introduced in Section $3$ is preferable to using only a single predicted time step.
\subsection{Connection Change}
\begin{figure}[th!]
     \centering
     \includegraphics[width=\linewidth]{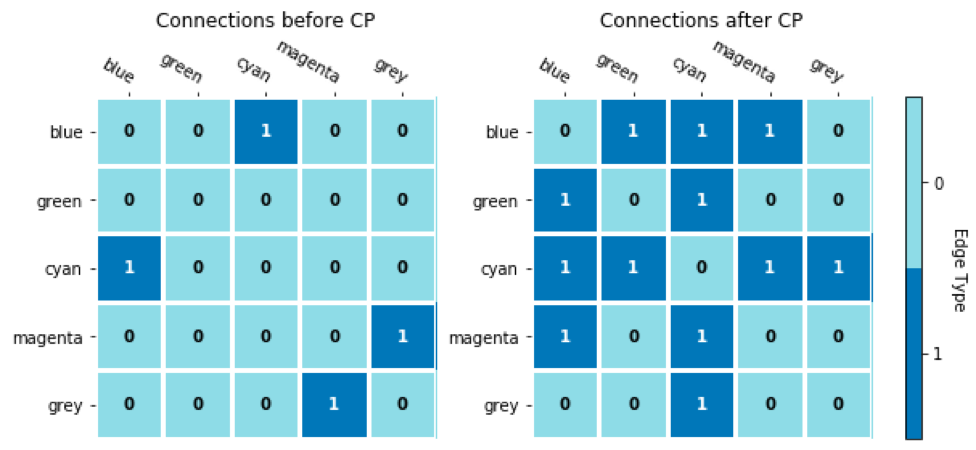}
     \caption{The connections before and after the change-point for the correlation change example.}
     \label{fig:con_edge}
\end{figure}
\begin{figure*}[th!]
     \centering
     \subfigure[Trajectory of Location Change \label{fig:syn_a}]{\includegraphics[width=\linewidth, height=6cm]{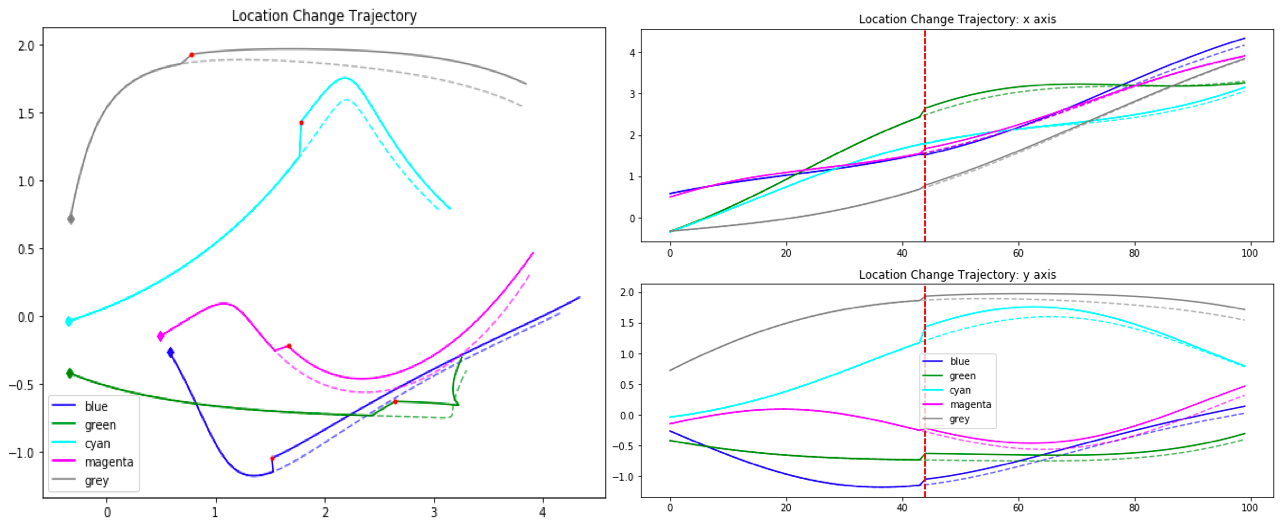}}
     \subfigure[Trajectory of Velocity Change \label{fig:syn_b}]{\includegraphics[width=\linewidth, height=6cm]{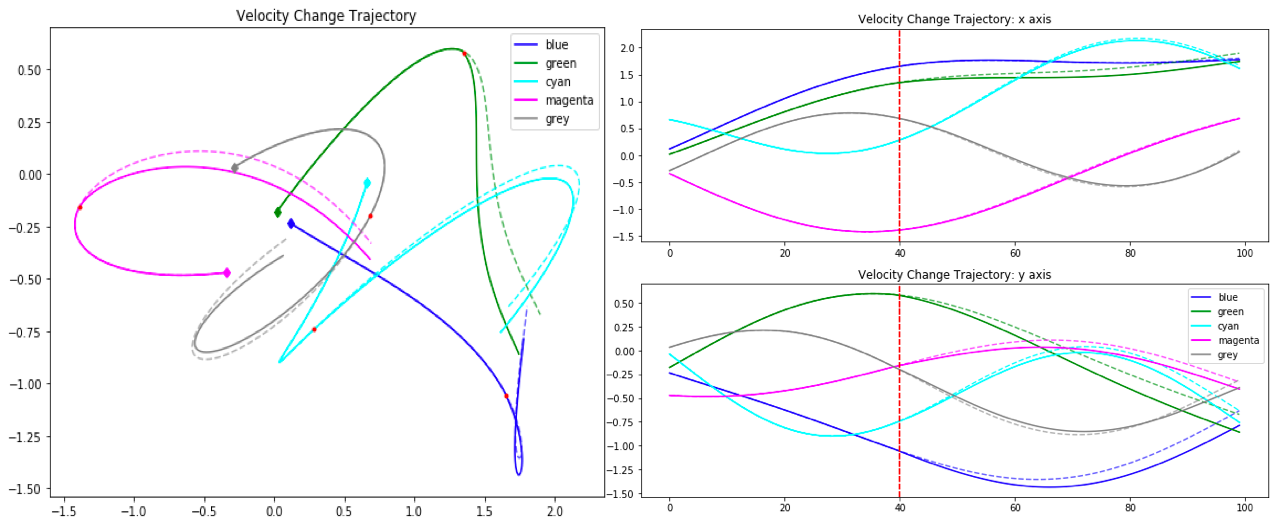}}
     \subfigure[Trajectory of Connection Change \label{fig:syn_c}]{\includegraphics[width=\linewidth, height=6cm]{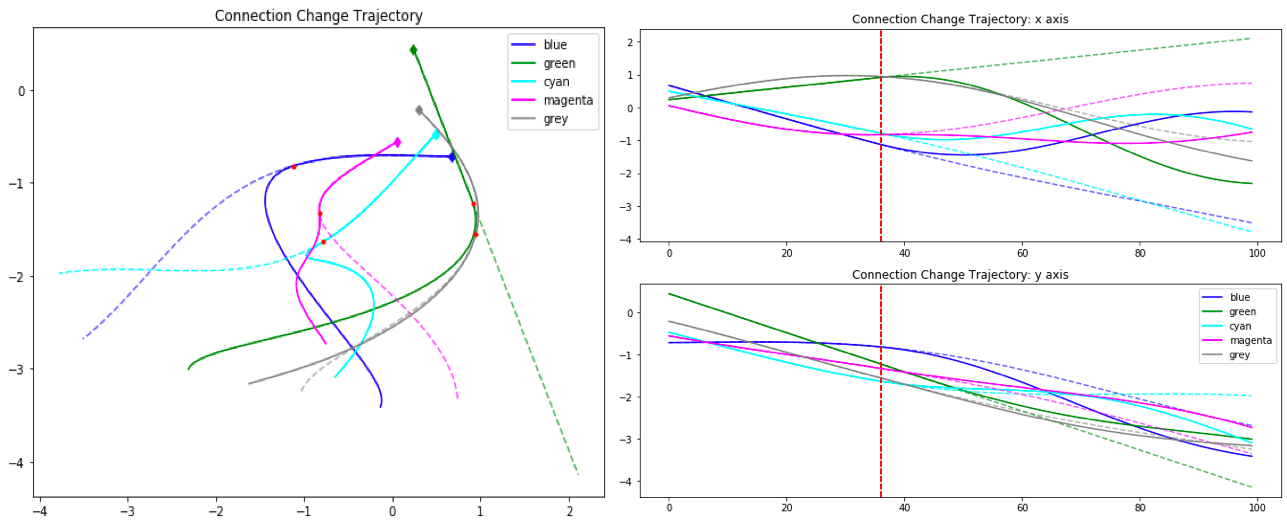}}
     \caption{The three figures show the trajectories of different types of change-points of $5$ particles (in 5 colors) connected by underlying springs. The figures on the left show the $2\mbox{-}D$ trajectories of the particles, and the figures on the right show the $x$ and $y$ axis of the trajectories separately. The dashed lines represent the expected trajectory if no change-point happened, and the solid lines are the actual observed trajectory.}
     \label{fig:syn_demo}
\end{figure*}

The connection change example is shown in Figure \ref{fig:syn_c}. Similar to the velocity change, the difference between the expected value and the observed value becomes more obvious over time. The Figure \ref{fig:con_edge} shows the underlying spring connections before and after the change-point. In particular, the \textit{green} particle was not connected with any other particles before the change-point, and it was connected with \textit{blue} and \textit{cyan} particles after the change-point. This altered the trajectory of the \textit{green} particle from a straight line to curved path. Detecting the spring re-connections change-points requires the modelling of dynamic correlations in multivariate time series, as our model did.

\section{Further Analysis on Synthetic Experiments}
In this section, we further describe our model training, baselines, and some other metrics on the synthetic dataset.

\subsection{Implementation Detail}
We perform a grid search for hyperparameters of the following values: the learning rate $l_r$ in $\{0.001, 0.005, 0.01, 0.05\}$, the hidden dimension size $d$ for the time series feature embeddings in $\{64, 128, 256\}$, and the number of levels of GNN or spatial transformer in $\{2, 3, 4\}$. We finally selected $l_r = 0.001$ using Adam optimizer, $d=64$ for transformers and $d=256$ for GNN. The level of GNN or spatial transformer is set as $2$, which is sufficient for our experiments. We used batch size of $32$ for temporal and spatial transformers and batch size of $128$ for GNN. 

We report the results for three encoder models: GNN\textsubscript{SEL}$+$RNN\textsubscript{TEL},
Trans\textsubscript{SEL} $+$ RNN\textsubscript{TEL} and 
GNN\textsubscript{SEL}$+$Trans\textsubscript{TEL}. Using both temporal and spatial transformer modules was hard to optimize, so we didn't include it as our model.

\subsection{Correlation prediction Accuracy}
Since the ground truth connections of springs ($\mathbf{A}$) are known in the synthetic dataset, we can calculate the accuracy of learnt correlation ($\hat{\mathbf{A}}$). The accuracy $p_{acc}$ is calculated by 
\[ p_{acc} = \frac{1}{T\times N \times (N-1)}\sum_{t=1, i<j}^{t\le T} \mathds{1}_{\{\hat{\mathbf{S}}^t_{i, j} = \mathbf{A}^t_{i,j}\}}  \]
$$
\hat{\mathbf{S}}^t_{i, j} = 
\begin{cases}
    1, & \text{if } \hat{\mathbf{A}}^t_{i, j} \ge 0.5 \\
    0, & \text{otherwise}
\end{cases}
$$
Where $\hat{\mathbf{S}}$ is the sampled categorical relation. $T$ is the number of time step and $N$ is the number of variables. For every pair of variables $i, j$, the function $\mathds{1}$ is an indicator function which outputs $1$ if $\hat{\mathbf{S}}^t_{i, j} = \mathbf{A}^t_{i,j}$, and $0$ otherwise.
\begin{table}[ht!]
\begin{center}
\begin{tabular}{c |c c c}
    \hline
    \bf {model} &  \bf location & \bf speed & \bf connection \\
    \hline
     \bf GNN\textsubscript{SE}+RNN\textsubscript{TE} & 96.07\% & 96.04\% & 90.45\% \\
     \bf Trans\textsubscript{SE}+RNN\textsubscript{TE} & \bf 97.79\% & 97.36\% & \bf 93.11\% \\
     \bf GNN\textsubscript{SE}+Trans\textsubscript{TE} & 97.49\% & \bf 97.47\% & 92.53\% \\
    \hline
\end{tabular}
\end{center}
\caption{\label{tab:acc-cls} The accuracy of predicted connections compared with ground truth connections in synthetic dataset.}
\end{table}

We observe that spatial and temporal transformers achieve better performance in the accuracy metrics, with Trans\textsubscript{SEL}$+$RNN\textsubscript{TEL} the being best on location and connection change, and nearly competitive as GNN\textsubscript{SEL}$+$Trans\textsubscript{TEL} on velocity changes. This result is consistent in the CPD task, such that Trans\textsubscript{SEL}$+$RNN\textsubscript{TEL} has the best score for separated predictions of independent changes and correlation changes.

\subsection{Correlation vs. Independent Change}
\begin{figure*}[th!]
     \centering
     \includegraphics[width=\linewidth]{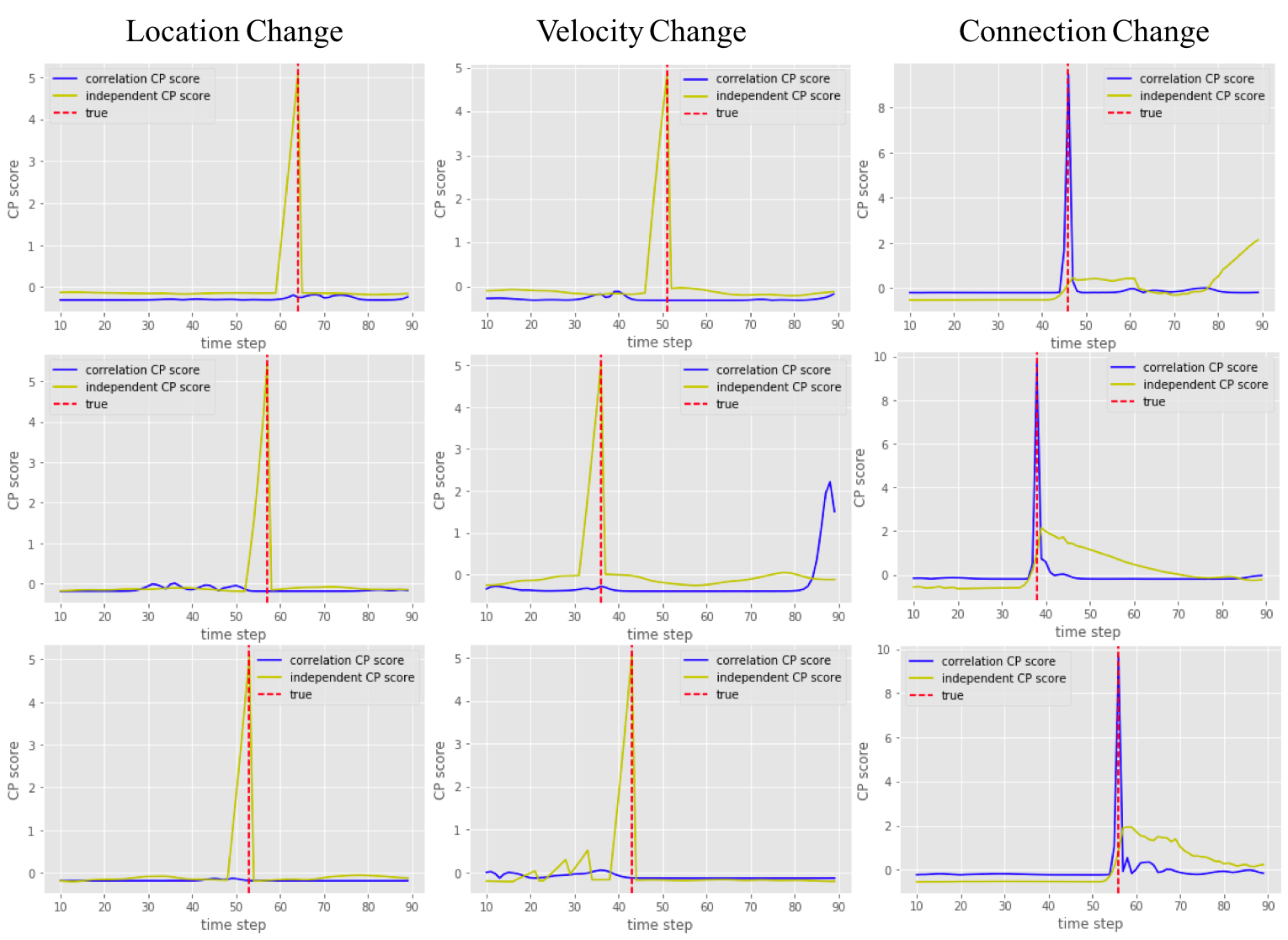}
     \caption{We show the correlation change-point score and the independent change-point score in the three types of change-points. The ground-truth change-point is labeled as red dashed line.}
     \label{fig:cls_example}
\end{figure*}

In the experiment, our model separately predicts the correlation change-point score by the correlation encoder, and the independent change-point score by the dynamics decoder. We also report the results when the two scores are separately evaluated. In Figure \ref{fig:cls_example}, we plot the two types of scores predicted by our model in the three types of changes, and the ground truth change-point label as the red dashed line. 

We observe that for location and velocity changes, the independent scores are peaked at the labeled change-point; For connection changes, the correlation scores are peaked at the labeled change-point. We conclude that model has the ability to separate the two types of change-points. 

\subsection{Change-point Type Classification}
\begin{figure}[th!]
     \centering
     \includegraphics[width=0.7\linewidth]{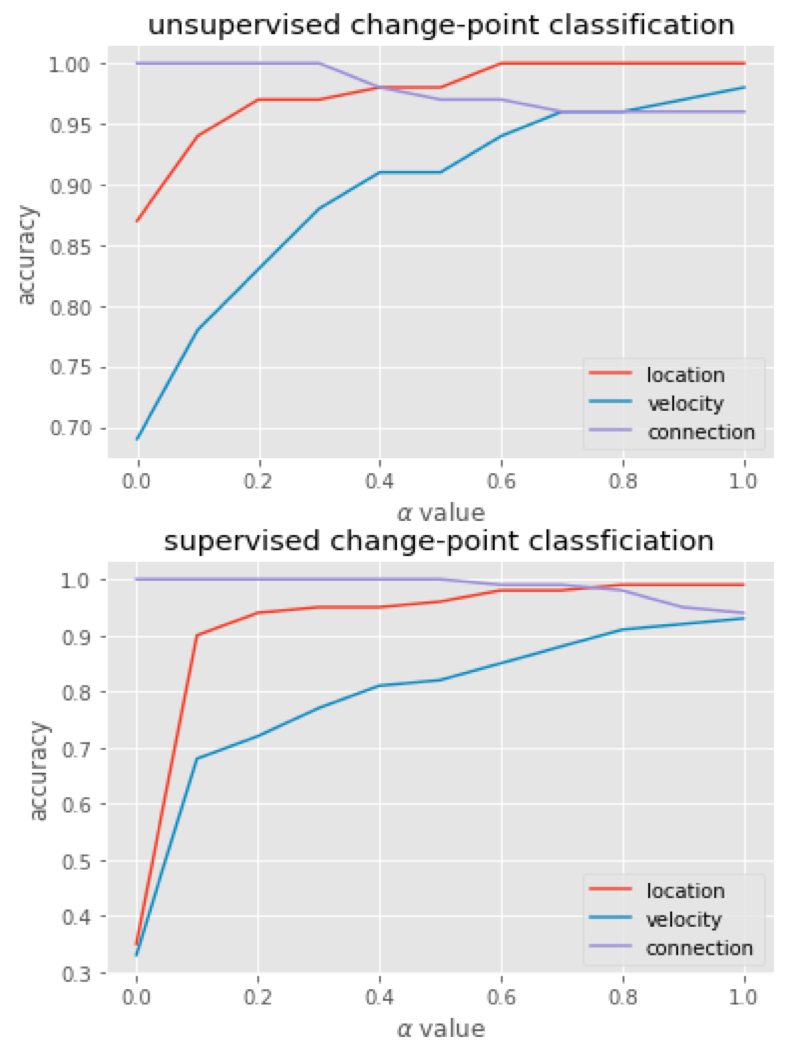}
     \caption{$\alpha$ value vs. the change-point type classification accuracy. $\alpha$ is in range $[0,1]$.}
     \label{fig:cls_alpha}
\end{figure}
In the change-type classification experiment in Section $5$, we propose to evaluate our model in both supervised setting and unsupervised setting. In the supervised setting, the time step of change-point is provided, and our goal is to predict the whether the change-point is resulted from an independent change or a correlation change. In the unsupervised setting, the time step of change-point not given, and we use the predicted change-point by our ensemble model instead. 

Our model separately predicts $s_r$, the correlation change-point score, and $s_d$, the independent change-point score. The change-point type is determined by $\operatorname{Norm}(s_r) - \alpha \operatorname{Norm}(s_d)$, such that
$$
\operatorname{Norm}(s_r) - \alpha \operatorname{Norm}(s_d)
\begin{cases}
     \ge \tau, & \text{ correlation change} \\
      < \tau, & \text{ independent change}
\end{cases}
$$
Where $\alpha$ is a hyperparameter and $\tau$ is a threshold. $\operatorname{Norm}$ is the mean-std normalization function.

In our study, we set $\tau = 0$ and visualize (in Figure \ref{fig:cls_alpha}) how $\alpha$ value affects the change-point type classification accuracy. The model we choose is GNN\textsubscript{SEL}$+$Trans\textsubscript{TEL}. We observe that if $\alpha$ is small, the correlation change-point score dominates, and connection changes are more accurately predicted. When $\alpha$ is large, the independent change-point score dominates, and the location and velocity changes are more accurately predicted. As a trade off between the two, we choose $\alpha=0.75$ in the experiment section. 

\subsection{Discussion on CPD}
\begin{figure}[th!]
     \centering
     \includegraphics[width=0.7\linewidth]{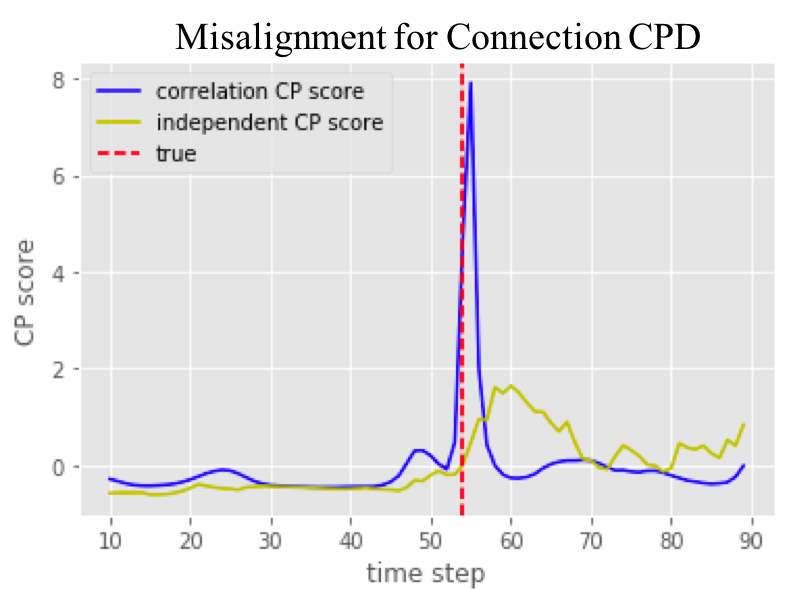}
     \caption{The peak of prediction is not aligned with the label, but very close due to a $1$-step delay.}
     \label{fig:misalign}
\end{figure}

In the experiments, we use two metrics: \textit{AUC-ROC}and \textit{TRI}. The AUC-ROC score was widely used in previous literatures \cite{li2015m, liu2013change, chang2019kernel} and TRI is what we proposed in this paper. The reason is that AUC-ROC treats each instance independently, but time-series data has a strong locality pattern. We do observe cases where the peak of the prediction is close to but not aligned with the labels, as shown in Figure \ref{fig:misalign}.

We observe that our model has the best performance in both metrics. For statistical and other deep learning baselines, we observe that they have similar AUC-ROC, but the statistical models are worse at TRI metrics. The reason is that statistical model concatenates a window sampled from training data at the start and end of each test case, to avoid the cold-start problem. However, this may lead the algorithm to give higher scores at the start or end of the test cases, resulting a larger gap from the labeled change-point.

\section{Robust CPD on Real Data}
In Table \ref{tab:real}, we report the performance of \ourmodel on the real-world PAMAP2 dataset. The multivariate time series includes $3$ variables and $10$ features. The variables are sensors on wrist, chest and ankle, and the features are temperature, 3D acceleration, gyroscope and magnetometer. The change-points are transitions between activities, such as walking, cycling, playing soccer. 

In the real-world scenario, the change-points are often resulted from a mixture of correlation change and independent change. We show the evaluated scores for the predicted correlation changes and independent changes of each model in Table \ref{tab:real}. We have two observations: 1) Each model capture similar trends for correlation changes and independent changes. There are more independent changes than correlation changes involved. 2) The ensemble of two reasons of change-points further boosts the performance, as the true reason of the change-point could include both of them.
\begin{table}[ht!]
\begin{center}
\setlength{\tabcolsep}{6pt}
\begin{tabular}{c c || c c c}
\hline
\bf {model} & \bf type & \bf AUC  & \bf TRI\\
\hline
 \bf \multirow{3}{*}{GNN\textsubscript{SE}+RNN\textsubscript{TE}} & rel & 0.6882 & 0.5972 \\
  & ind & 0.7850 & 0.7088 \\
 & ens & 0.7868 & 0.7574 \\
\hline
 \bf \multirow{3}{*}{Trans\textsubscript{SE}+RNN\textsubscript{TE}} & rel & 0.6538  & 0.4118 \\
& ind & 0.7722 & 0.7360 \\
& ens & 0.7903  & 0.7750 \\
\hline
 \bf \multirow{3}{*}{GNN\textsubscript{SE}+Trans\textsubscript{TE}} & rel & 0.6715  & 0.4013 \\
& ind &  0.7787  & 0.7102 \\
& ens & \bf 0.8277  & \bf 0.8020 \\
\hline
\end{tabular}
\end{center}
\caption{\label{tab:real} The performance of our \ourmodel on a real-world PAMAP2 dataset for CPD. We include the scores of independent changes and correlation changes as well. 
}
\end{table}

\end{document}